# *Prediction of Brent crude oil price based on LSTM model under the background of low-carbon transition*

**Yuwen Zhao[1,a,*], Baojun Hu[2,b], Sizhe Wang[3,c]**

[1]University of Sydney, Australia
[2]Information Systems & Operations Management, Warrington college of business, university of florida, USA
[3]Lehigh University, Dallas, TX, USA
a. tauriel.z@outlook.com, b. corinthpersonal@163.com, c. siw422@lehigh.edu

*Abstract:* In the field of global energy and environment, crude oil is an important strategic resource, and its price fluctuation has a far-reaching impact on the global economy, financial market and the process of low-carbon development. In recent years, with the gradual promotion of green energy transformation and low-carbon development in various countries, the dynamics of crude oil market have become more complicated and changeable. The price of crude oil is not only influenced by traditional factors such as supply and demand, geopolitical conflict and production technology, but also faces the challenges of energy policy transformation, carbon emission control and new energy technology development. This diversified driving factor makes the prediction of crude oil price not only very important in economic decision-making and energy planning, but also a key issue in financial markets.In this paper, the spot price data of European Brent crude oil provided by us energy information administration are selected, and a deep learning model with three layers of LSTM units is constructed to predict the crude oil price in the next few days. The results show that the LSTM model performs well in capturing the overall price trend, although there is some deviation during the period of sharp price fluctuation. The research in this paper not only verifies the applicability of LSTM model in energy market forecasting, but also provides data support for policy makers and investors when facing the uncertainty of crude oil price.

*Keywords:* LSTM model; Low Carbon Economy; Crude oil futures; price projection

## 1. Introduction

The volatility of global energy market has always been an important factor affecting international economic activities, especially in the context of accelerating global energy transformation and low-carbon development, the impact of crude oil price fluctuations on the global economic system has become more and more prominent. As an important driving force of the international economy, the change of crude oil price not only affects the economic structure of oil exporting and importing countries, but also has a far-reaching impact on the operation of major economies around the world through commodity markets, financial markets and global supply chains. The fluctuation of crude oil price is closely related to many factors such as world economic cycle, financial market fluctuation, exchange rate change, monetary policy, international trade relations and geopolitical events. Since 2000, the international crude oil price has experienced several significant fluctuations. In 2008, the global financial crisis led to a sharp drop in crude oil prices.

At the same time, the global energy structure is undergoing profound changes with the promotion of low-carbon development goals. The rapid development of renewable energy and the improvement of energy efficiency are gradually reducing the dependence on fossil fuels, squeezing the demand of traditional energy market. As one of the main actions under the framework of Paris Agreement, many economies in the world have put forward the goal of carbon neutrality, which not only promotes the investment growth in the field of clean energy, but also brings great uncertainty in the fossil fuel market, thus further aggravating the volatility of crude oil prices.

In this context, accurate prediction of crude oil prices has become an urgent problem for policy makers and financial market participants. Traditional forecasting methods mostly rely on the analysis of macroeconomic indicators and supply and demand fundamentals. However, with the increasing data scale and increasingly complex market environment, the limitations of traditional methods are gradually emerging. Based on the in-depth analysis of historical data and the application of LSTM model, this paper discusses the performance of this model in forecasting Brent crude oil price, evaluates its applicability in energy market analysis, and provides new analytical ideas and methods and tools for the complexity of crude oil market under the background of low-carbon development.

## 2. Literature Review

The research of crude oil price forecast has always been a hot topic in economics, finance and energy market research. Many scholars have conducted in-depth research from different angles and methods. In this regard, deep learning models such as LSTM have been widely used in crude oil price forecasting. Jiang Feng et al. put forward a crude oil futures price forecasting model based on multi-source and multi-task automatic encoder, which improved the accuracy of forecasting by effectively using multi-source data and multi-task learning mechanism [1]. Pan Shaowei et al. and Yu Wenmei et al. also used LSTM to forecast crude oil prices, and achieved good forecasting results [2].

At the same time, some researchers try to combine LSTM with other forecasting models to further improve the accuracy of forecasting. Liao Jingwen proposed an international crude oil price forecasting model based on VMD-LSTM-ELMAN [3]. By introducing variational modal decomposition (VMD) and ELMAN neural network, the forecasting accuracy was significantly improved. Li Yang et al. put forward a study on crude oil production time series prediction based on ARIMA-LSTM [4]. This model combines ARIMA model and LSTM model, and the prediction effect is remarkable.

In addition to the deep learning model, researchers also explore the prediction of crude oil prices from different perspectives. Liu Wenjing et al. carried out the forecast and analysis of crude oil price according to the influence of multiple factors [5]. Lin Yu et al. proposed a crude oil futures price forecasting model based on error correction and deep reinforcement learning, which made up for the shortcomings of the traditional forecasting model [6]. Lv Chengshuang et al. put forward a research on international crude oil price prediction based on CATTSTS model, and improved the prediction accuracy by introducing the periodic trend prediction method [7].

In addition, some researchers try to combine the deep learning model with other methods and put forward a new prediction model. For example, Zhao Goya et al. proposed a crude oil futures next-day price forecasting model based on TN-LP-LSTM-SVM hybrid model, which effectively improved the accuracy of forecasting [8].

Numerous studies have shown that deep learning models, especially LSTM, can not only be used for crude oil price forecasting alone, but also be effectively combined with other models to improve the accuracy of forecasting [9-11]. At the same time, multi-factor analysis and new forecasting methods also provide a new perspective for crude oil price forecasting. These studies provide valuable reference for understanding and forecasting the price of crude oil in this paper.

## 3. Data and Analysis

### 3.1. Data Introduction

The trend of crude oil price is complex and changeable, which is influenced by multiple factors, such as macroeconomic factors, geopolitical events, supply-demand relationship and market expectations. In order to effectively predict the price fluctuation caused by these complex factors, this paper selects the spot price data of European Brent crude oil provided by us energy information administration (EIA) as the research object. Brent crude oil price is widely regarded as the benchmark of international crude oil price because of its importance in the global oil market. Its price data not only reflects the global supply and demand dynamics, but also includes the market's expectation of future prices and the influence of various economic and political factors.

The data set used in this paper covers the daily spot price of Brent crude oil from January 1, 2018 to September 30, 2019, and the unit is USD/barrel. In order to improve the prediction performance of the model, the original data is transformed logarithmically, which makes the price data more stable in the time series and helps to reduce the errors in model training.

### 3.2. Model Introduction

In the aspect of model selection, this paper uses LSTM (long-term and short-term memory network) model to predict the time series of crude oil prices. The LSTM model is a special recurrent neural network (RNN), which can effectively capture the long-term dependence in time series by introducing a Memory Cell. The traditional RNN is prone to the problem of gradient disappearance or gradient explosion when processing long series data, which makes it difficult for the model to remember the early information. LSTM can effectively solve these problems through the mechanisms of "forgetting gate", "input gate" and "output gate" in its internal structure, thus performing well in dealing with long-term dependent data.

Based on the trained LSTM model, this paper forecasts the price of Brent crude oil in the next few days. By comparing with the actual data, this paper evaluates the prediction accuracy of the model in detail, and discusses the advantages and limitations of LSTM model in dealing with the complex fluctuation of crude oil prices. At the same time, the forecast results of this paper not only provide valuable reference for the future trend of crude oil prices, but also provide data support for participants in the energy market when making investment decisions.

### 3.3. Data description analysis

This paper extracts the data from 2000 to 2019, resamples the price data on a monthly basis, and generates a time series of monthly average prices. Furthermore, Figure 1 of this time series is drawn to show the long-term trend and fluctuation of Brent crude oil price during this time period.

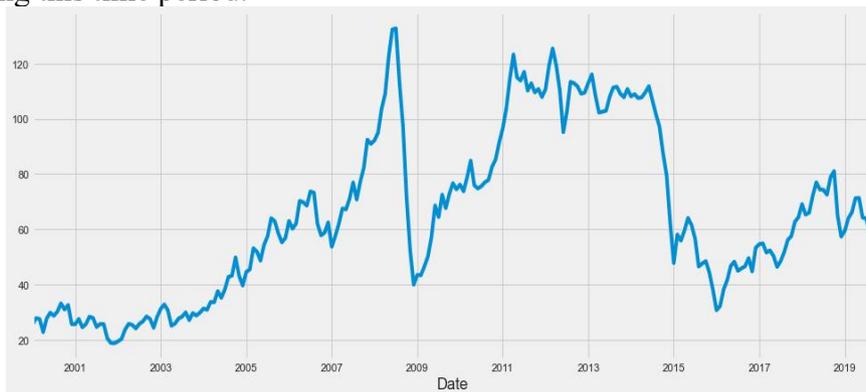

Figure 1: Brent crude oil price time series figure

The geometric Brownian motion model is used to simulate the future price of Brent crude oil, and 50 potential price paths are generated, and the average values of these paths are calculated to predict the future price trend. Figure 2 shows the historical price from January 1, 2018, as well as the simulated potential price paths and their average predicted prices, so as to show the potential fluctuation range and possible average trend of future prices.

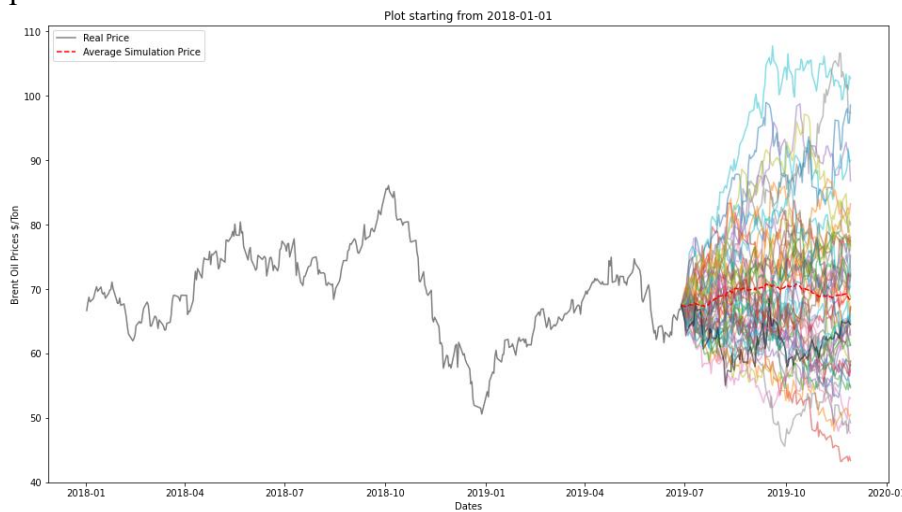

Figure 2: Simulation figure of geometric Brownian motion model

### 4. LSTM model analysis

In order to ensure that the training process of the model is more stable and effective, the data set is normalized to between 0 and 1. Then, the data set is divided into 70% training set and 30% testing set to ensure that the model can be tested on new data. Next, the sliding window method is used, and each input sample is set to contain historical data of 90 time steps, so that the LSTM model can predict future values based on past data.

When constructing the LSTM model, this paper designs three layers of LSTM elements, each layer contains 60 elements, and adds a Dropout layer after each layer to reduce the risk of over-fitting. Adam optimizer is used in the model training process, and the verification loss is optimized by reducing the learning rate. After the training, the prediction results of the model on the training set and the test set are denormalized, and the mean absolute error (MAE) and root mean square error (RMSE) are calculated to evaluate the accuracy of the model.

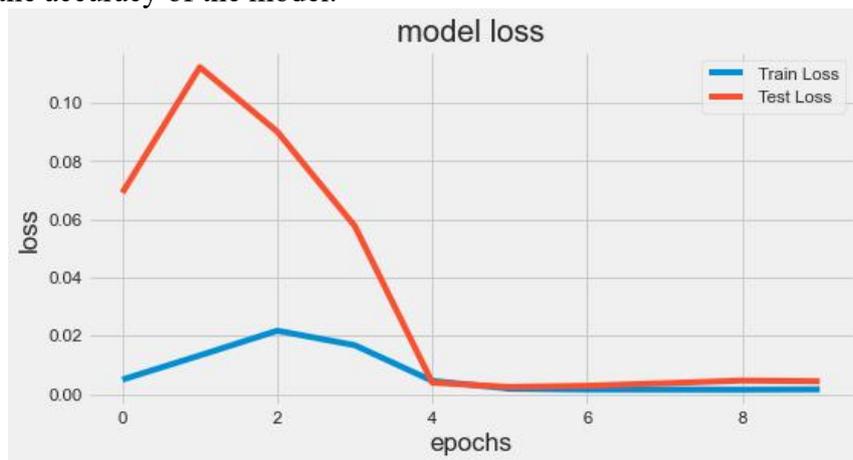

Figure 3: LSTM Model loss curve

Figure 3 shows the loss changes of the model during training. Figure 3 shows that the training loss decreases steadily and tends to be stable after the fifth epoch, while the test loss rises sharply at the beginning, decreases rapidly after reaching the peak, and finally is almost consistent with the training loss. This shows that the model may be over-fitted in the initial stage, but after further training, the model is more stable and has better generalization ability.

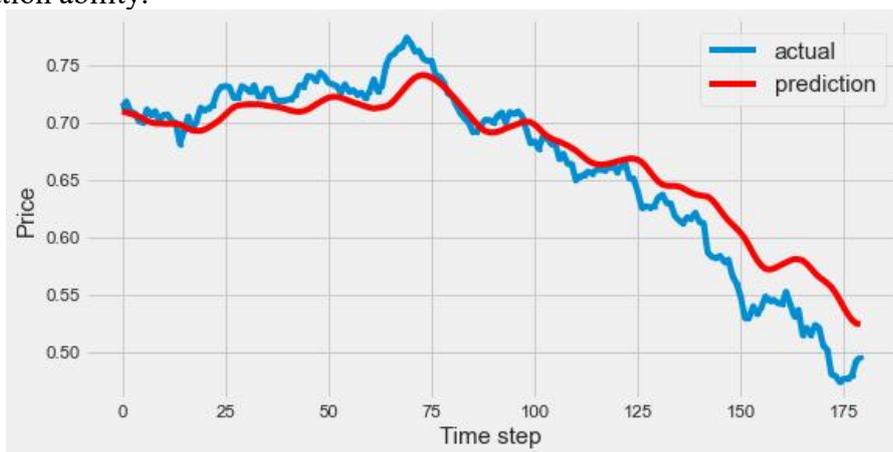

Figure 4: LSTM prediction curve

Figure 4 shows the difference between the predicted results of the LSTM model and the actual data. In the chart, the horizontal axis represents time step, the vertical axis represents price, the blue line represents actual price, and the red line represents the predicted price of the model. By comparing the trends of these two lines, the prediction accuracy of the model can be evaluated intuitively. The chart shows the prediction performance of the model in the first 180 time steps, which is helpful to understand the fitting degree of the model to time series data.

Figure 4 shows that although the overall trend of the predicted value (red line) of the model is close to the actual value (blue line), there are still some deviations, especially in the areas where prices fluctuate. On the whole, the model can capture the downward trend of prices, but it can't accurately track the actual price changes in some time periods.

## 5. Conclusion

In this paper, the LSTM model is used to predict the price of Brent crude oil, which shows the potential of deep learning in capturing the long-term dependence of crude oil market and its price trend. The results show that the LSTM model can effectively predict the trend of crude oil price in most cases, especially in the period of relatively stable market, although the prediction accuracy decreases during the period of large price fluctuation. This discovery provides a new tool and perspective for policy makers and investors in the face of market uncertainty.

However, considering the acceleration of global low-carbon transformation and the constant change of crude oil market, traditional forecasting methods are facing challenges. The introduction of LSTM model not only improves the adaptability to complex market environment, but also provides a foundation for further research. Future work can explore the combination of LSTM and other forecasting technologies to enhance the performance of the model in extreme market conditions, and at the same time, we can consider introducing more factors that affect the price of crude oil to build a more comprehensive and accurate forecasting model.